\documentclass[10pt,twocolumn,letterpaper]{article}

\usepackage[pagenumbers]{iccv} % To force page numbers, e.g. for an arXiv version
\usepackage{amsmath}
\usepackage{bm}
\usepackage{tabularx}
\usepackage{graphicx}
\usepackage{multirow}
\usepackage{adjustbox}
\usepackage{float}
\usepackage{xr}
\usepackage{mathrsfs}
\usepackage{booktabs}
\usepackage{tensor}
\definecolor{tabfirst}{rgb}{1, 0.7, 0.7} % red
\definecolor{tabsecond}{rgb}{1, 0.85, 0.7} % orange
\definecolor{tabthird}{rgb}{1, 1, 0.7} % yellow
\usepackage{amsmath}
\usepackage{wrapfig}
\definecolor{iccvblue}{rgb}{0.21,0.49,0.74}
\usepackage[pagebackref,breaklinks,colorlinks,allcolors=iccvblue]{hyperref}

\def\paperID{3908} % *** Enter the Paper ID here
\def\confName{ICCV}
\def\confYear{2025}

\title{GaRe: Relightable 3D Gaussian Splatting for Outdoor Scenes from Unconstrained Photo Collections}

\author{Haiyang Bai$^{1}$ \quad 
Jiaqi Zhu$^{1}$ \quad
Songru Jiang$^{1}$ \quad
Wei Huang$^{1}$ \quad
Tao Lu$^{2}$ \\
Yuanqi Li$^{1}$ \quad
Jie Guo$^{1}$ \quad
Runze Fu$^{3}$ \quad
Yanwen Guo$^{1}$ \quad
Lijun Chen$^{1,*}$ \\ \\
$^{1}$Nanjing University \quad
$^{2}$Brown University \quad
$^{3}$JSTI Group \\ \\
\url{https://baihyyut.github.io/GaRe/}
}
\small

\begin{document}
% \maketitle

\twocolumn[{%
\renewcommand\twocolumn[1][]{#11}%
\maketitle
\begin{center}
    \centering
    \captionsetup{hypcap=false}
    \includegraphics[width=1.0\textwidth]{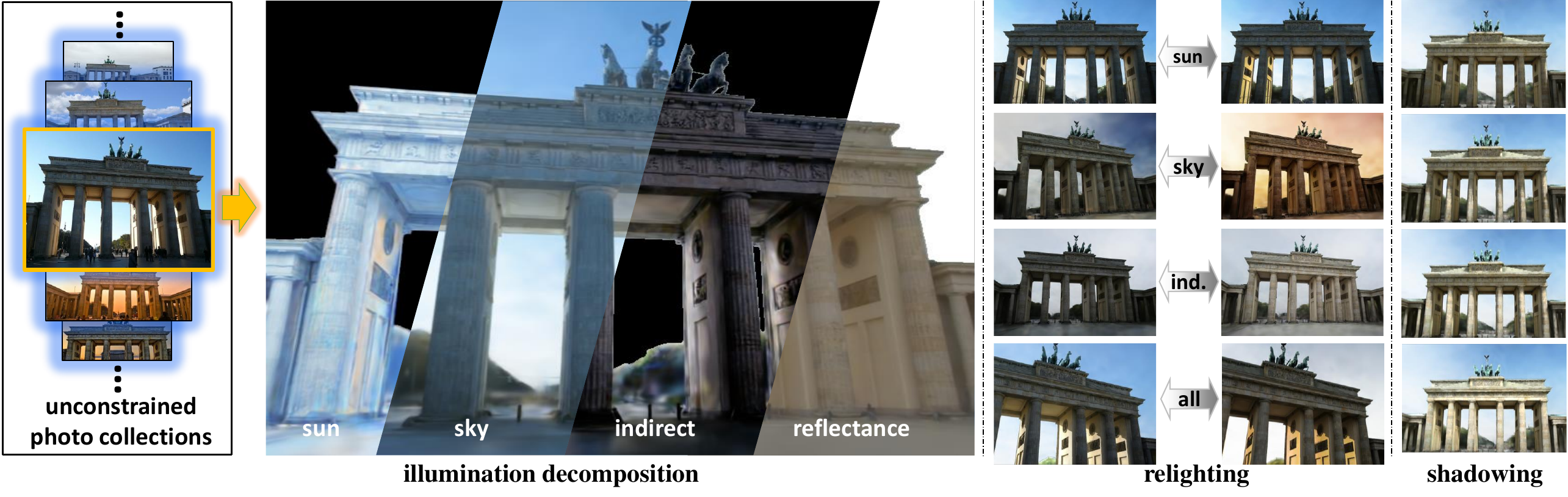}
    \captionof{figure}{Building on the principles of intrinsic image decomposition, our relighting framework, termed $ \textbf {GaRe}$, accurately separates reflectance and multiple physically interpretable shading components—sunlight, sky radiance, and indirect lighting—from unconstrained outdoor photo collections. This meticulous decomposition transforms these components into versatile lighting assets, enabling precise and highly adaptive relighting.  Each component can be manipulated independently or cohesively integrated (as represented by "all" in the figure) with shadows, offering unparalleled flexibility to achieve diverse and sophisticated relighting effects.}
\captionsetup{hypcap=true}
\label{fig:teaser}
\end{center}%
}]

\footnote{* Corresponding author.}

\begin{abstract}

% We propose a Gaussian-splatting-based framework for outdoor relighting that leverages intrinsic image decomposition to effectively integrate sunlight, sky radiance, and indirect lighting from unconstrained photo collections. The proposed method incorporates three key elements, enabling real-time precise shading manipulation and the generation of dynamic shadow effects. First, we extract sun visibility by applying binary clustering to the residuals between globally illuminated images and their ambient-only rendered counterparts, ensuring accurate separation of direct sunlight effects. Second, region-based supervision is used to jointly optimize per-Gaussian illumination attributes and per-image shading embeddings, resulting in physically interpretable shading components. Meanwhile, we introduce a novel structural consistency loss to improve relighting consistency. Third, we apply point-based ray tracing to perform ray-dependent visibility queries, enabling realistic shadow simulation. For real-time relighting, ray-traced shadows are baked into Gaussian representations parameterized by local visibility features. This allows all illumination components and shadow conditions to be rendered efficiently in a single pass. Extensive experiments demonstrate that our framework synthesizes novel views with competitive fidelity against state-of-the-art relighting solutions and produces more natural and multifaceted illumination and shadow effects.

We propose a 3D Gaussian splatting-based framework for outdoor relighting that leverages intrinsic image decomposition to precisely integrate sunlight, sky radiance, and indirect lighting from unconstrained photo collections. Unlike prior methods that compress the per-image global illumination into a single latent vector, our approach enables simultaneously diverse shading manipulation and the generation of dynamic shadow effects. This is achieved through three key innovations: (1) a residual-based sun visibility extraction method to accurately separate direct sunlight effects, (2) a region-based supervision framework with a structural consistency loss for physically interpretable and coherent illumination decomposition, and (3) a ray-tracing-based technique for realistic shadow simulation. Extensive experiments demonstrate that our framework synthesizes novel views with competitive fidelity against state-of-the-art relighting solutions and produces more natural and multifaceted illumination and shadow effects. 

% Project page: \url{https://baihyyut.github.io/GaRe.github.io/}.

% Project page: \url{https://tex.nju.edu.cn/}.

\end{abstract}
\section{Introduction}
\label{sec:intro}

Realistic relighting is a fundamental task at the intersection of computer vision and graphics, with applications ranging from virtual and augmented reality (VR/AR) to cinematic visual effects. 
Traditional physics-based rendering (PBR) methods~\cite{pharr2023physically, tsirikoglou2017procedural, haro2001real} leverage ray/path tracing algorithms to simulate light transport, facilitating precise global illumination and shadow computations. However, their reliance on detailed geometries poses significant challenges for deployment in unconstrained outdoor scenarios.
In contrast, volumetric representations~\cite{nerf, toschi2023relight, martin2021nerf, chen2022hallucinated, fridovich2023k} embrace a learning-based paradigm, reformulating relighting task as the mapping of scene illumination from captured images onto radiance fields. While more flexible, these methods are hindered by high computational overhead, making them less suitable for real-time applications.

Recently, 3D Gaussian Splatting (3DGS)~\cite{kerbl20233d} introduced an explicit
Gaussian-based scene representation, enabling photorealistic rendering at real-time speed.
Subsequently, further research~\cite{gao2023relightable, liang2024gs, jiang2024gaussianshader} in relighting has shown encouraging outcomes. 
Nonetheless, they mainly focus on encoding per-image illumination into a single latent vector~\cite{zhang2025gaussian, kulhanek2024wildgaussians}, which facilitates only global appearance adjustments while limiting the exploration of diverse outdoor shading effects, such as sunlight, sky shading, and other intricate lighting interactions. To address this limitation, intrinsic image decomposition~\cite{duchene2015multi, laffont2012rich} offers a potential solution by separating reflectance and diverse illumination elements from outdoor images, allowing for independent editing and recomposition while maintaining consistent lighting and realistic shadows. However, they lack the flexibility to change the scene’s capture time or adjust weather conditions, as such changes often lead to inaccuracies in illumination and shadow predictions.

In this paper, we propose \textbf{GaRe}, a \textbf{Ga}ussian splatting-based outdoor real-time \textbf{Re}lighting framework, enabling simultaneously diverse shading manipulation and the generation of shadow effects.
Inspired by intrinsic image decomposition, we explicitly integrate sunlight, sky radiance, and indirect lighting from unconstrained photo collections into volumetric Gaussians. 
%
% The framework comprises three core elements. 
%
To separate the sunlight effects, we first extract a sun visibility map via binary clustering on the residuals between globally illuminated images and ambient-only rendered counterparts, where the residuals delineate the regions of illumination and shadow.
Second, we employ region-based supervision to jointly optimize per-Gaussian illumination features and per-image shading embeddings, effectively decomposing global illumination into physically interpretable shading components. 
To ensure coherent relighting, we introduce a structural consistency loss that propagates locally optimized shading across the entire color domain. Finally, we utilize point-based ray tracing for visibility queries, enabling realistic shadow simulation. 
Furthermore, we bake ray-traced shadows into Gaussian local visibility features, enabling real-time rendering of illumination components and shadows in a single pass.
As illustrated in Fig.~\ref{fig:teaser}, our relighting produces more natural and multifaceted illumination and shadow effects.

To summarize, our key contributions are as follows:
\begin{itemize}
\item We propose a novel visibility extraction method by applying binary clustering to the residuals between globally illuminated images and ambient-only rendered counterparts, achieving precise separation of direct sunlight effects.

\item We introduce a region-based supervision framework with a structural consistency loss for physically interpretable and coherent illumination decomposition.

\item We integrate ray tracing into Gaussian representations and design a shadow baking process, achieving real-time relighting with single-pass efficiency.

\end{itemize}
 
\begin{figure*}[t]
  \centering
  \includegraphics[width=1.0 \textwidth]{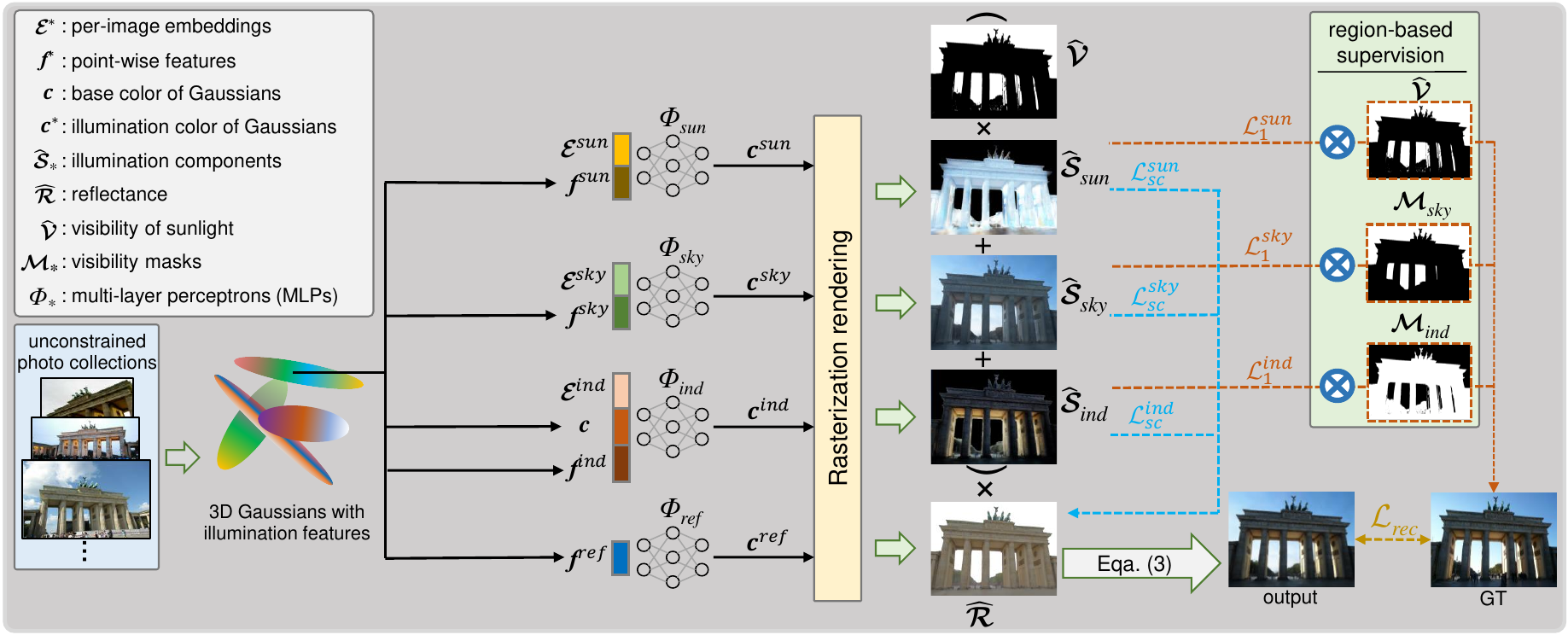}
  \caption{Overview of illumination decomposition. The GaRe framework accurately decomposes global illumination into reflectance and distinct shading components by associating each Gaussian with shading-related features $ \{ {\bm f^{ref}}, {\bm f^{sun}}, {\bm f^{sky}}, {\bm f^{ind}} \} $, which are jointly decoded with per-image embeddings $ \{ \mathcal{E}^{sun},  \mathcal{E}^{sky}, \mathcal{E}^{ind} \} $ to estimate corresponding spartial shading and reflectance colors $ \{ {\bm c^{sun}}, {\bm c^{sky}}, {\bm c^{ind}}, {\bm c^{ref}} \} $.  To enhance the physical interpretability, a region-supervised loss term $ \mathcal{L}_1^* $ is employed to optimize the rendered sun, sky, and indirect shading components $ \{ \hat{\mathcal{S}}^{sun}, \hat{\mathcal{S}}^{sky}, \hat{\mathcal{S}}^{ind} \} $ by segmenting illumination regions using visibility masks $ \{ \hat {\mathcal{V}}, \mathcal{M}_{sky}, \mathcal{M}_{ind} \} $, while a novel structural consistency loss $ \mathcal{L}_{sc}^* $ is introduced to ensure the global consistency and physical plausibility of the illumination during the relighting process.}
  \label{fig:framework}
\end{figure*}

\section{Related Work}
\label{sec:related_work}
% The evolution of outdoor scene relighting techniques spans multiple paradigms, from traditional methods to neural rendering approaches. A significant challenge across these paradigms is that existing outdoor scene datasets typically suffer from limited image resolution and lack explicit light source annotations. Given these constraints, we approach the relighting task from a more generalized perspective, where light source information is assumed to be unavailable. In the following sections, we present a chronological review of the most pertinent works in this domain. 

\subsection{Traditional Relighting}
Traditional relighting approaches can be broadly categorized into physics-based methods and inverse rendering techniques. Traditional physics-based graphics methods primarily employ ray tracing~\cite{whitted2005improved, wald2003realtime, majercik2019dynamic} or radiosity algorithms~\cite{hanrahan1991rapid, hachisuka2008progressive, sloan2023precomputed} to render scenes freely, which naturally enables relighting. For effectiveness, \cite{wang2006gpubased, ng2003allfrequence, kristensen2005precomputed} leverage wavelets and spherical harmonics with reflectance distribution functions to simulate lighting interactions. On the other hand, \cite{nayar1991shape, marschner1998inverse} establish core principles for inverse rendering. Following image-based inverse rendering techniques~\cite{bell2014intrinsic, laffont2012coherent} emerge to decompose images into reflectance and shading components at the pixel level.  Recent approaches ~\cite{li2020inverse, nam2018svbrdf, wu2016simultaneuos, xia2016recovering} commit to jointly optimize the geometry and reflectance. Despite their respective merits, these approaches face significant limitations: physics-based methods require expensive explicit 3D meshes and accurate material properties, while inverse rendering techniques remain confined to controlled indoor environments and small objects with both computational complexity and reconstruction errors significantly increasing with scene scale.

\subsection{Radiance Field-Based Relighting}
Radiance fields methods, particularly NeRF~\cite{nerf, barron2021mip} and 3DGS~\cite{kerbl20233d, lu2024scaffold}, provide an end-to-end approach that jointly addresses geometry and additional attributes (\eg normals, depth, reflectance). Building upon NeRF, NeRD~\cite{boss2021nerd} and NeRFactor~\cite{zhang2021nerfactor} integrate physical attributes to recover materials and diverse lighting, while~\cite{zeng2023relighting} introduces shadow and highlight hints to enhance relighting quality. Some research~\cite{Ye2023IntrinsicNeRF, nerv2021} addresses intrinsic decomposition, while \cite{kuang2022neroic, guo2022nerfren} extend these capabilities to handle environment map estimation and complex scene lighting. However, these methods are heavily limited by computational demands.
Recent works have also extended 3DGS~\cite{kerbl20233d} for relighting applications: Gaussianshader~\cite{jiang2024gaussianshader} aims to simplify shading new environment maps to the scene, while other comprehensive relighting systems~\cite{gao2023relightable, shi2023gir,bi2024gs3} integrate relightable attributes to gaussian proxies and employ ray tracing visibility to resolve occlusion problems. Although these methods yield impressive results for object-level relighting under controlled conditions, they remain limited in managing complex outdoor scenes with unconstrained conditions.

\subsection{Relighting for Outdoor}
In the context of general scenarios (\eg outdoor scenes), 3DGS~\cite{kerbl20233d} and subsequent relighting works~\cite{liang2024gs,gao2023relightable, shi2023gir} focus on multi-view datasets with fixed illumination. While effective, such constrained datasets require significant acquisition effort and fail to address the complexity of real-world outdoor scenes. Unstructured photo collections provide greater accessibility but introduce challenges like variable illumination and dynamic objects (\eg pedestrians, and vehicles). Several pioneering works~\cite{philip2019multi, griffiths2022outcast} investigate scene relighting using single real-world images, which subsequently inspire recent advancements~\cite{martin2021nerf, chen2022hallucinated,zhang2025gaussian, kulhanek2024wildgaussians} that demonstrate remarkable capabilities in novel view synthesis from such collections. Nonetheless, these methods do not address the critical challenge of relighting. Drawing inspiration from recent progress in intrinsic decomposition and inverse rendering~\cite{philip2019multi, li2020inverse, li2022neulighting, zhang2022invrender, li2023multi, Jin2023TensoIR}, we present a novel framework that not only achieves state-of-the-art scene modeling accuracy but also enables robust illumination decomposition and realistic relighting from unconstrained outdoor photo collections.

\section{Our Method: GaRe}

\subsection{Framework Overview}

Given a collection of unconstrained outdoor images captured under varying temporal and weather conditions, we model the scene using Lambertian surfaces~\cite{oren1993diffuse}, where the observed appearance is represented as the product of the incident illuminations and the surface reflectance $ {\mathcal R} $~\cite{duchene2015multi, laffont2012rich}. 
To facilitate relighting, the incoming radiance is further decomposed into three distinct shading components: direct radiance from the sun $ {\mathcal S}_{sun} $, diffuse radiance from the sky $ {\mathcal S}_{sky} $, and indirect radiance resulting from global illumination $ {\mathcal S}_{ind} $. 
The final rendered image $ {\mathcal I} $ is thus formulated as: 
\begin{equation}
\label{Eqn:intrinsic_image}
    {\mathcal I} = ( {\mathcal V} \cdot {\mathcal S}_{sun} + {\mathcal S}_{sky} + {\mathcal S}_{ind}) \cdot {\mathcal R},
\end{equation}
where $ {\mathcal V} $ denotes the sun visibility map that identifies pixels directly lit by the sun, thereby enabling explicit shadow control. Our framework follows a three-step paradigm: (1) extraction of sun visibility (Sec. \ref{sec:visibility_extraction}), (2) decomposition of shading components $ \mathcal S_\ast $  and reflectance $ \mathcal R $ (Sec. \ref{sec:illumination_decomposition}), and (3) shadow modeling via ray-tracing (Sec.~\ref{sec:gaussian_visibility}).
%
% Firstly, we pre-train a standard Gaussian model with ambient-only scene images, the visibility map $ \mathcal V $ is then obtained through pixel-wise clustering on the residual map between ambient-only predictions and ground truth images with global illumination (Sec.\ref{sec:visibility_extraction}).
% %%
% Secondly, the full canvas is separated by a set of disjoint region masks $ \mathcal M_\ast $ w.r.t. each shading component $ \mathcal S_\ast $, ensuring them physically interpretable (Sec. \ref{sec:illumination_decomposition}). After that, the region-based supervision is then applied to learn the responsible illumination features in our extended Gaussian model.
% %%
% Finally, we develop a ray-tracing-based method for vivid shadow effects modeling. We bake the raw ray-traced shadows into Gaussian representation, parameterized by learnable visibility features, to enable efficient rendering of the illumination components and ray-dependent shadows all in a single pass (Sec.~\ref{sec:gaussian_visibility}).

% \vspace{-0.3cm}
\subsection{Residual-based Visibility Extraction}
\label{sec:visibility_extraction}
% Accurate isolation of sunlight hinges on reliably distinguishing sun visibility within a view. Previous methods~\cite{li2022neulighting, bi2024gs3}, which leverage ray-surface intersection detection, encounter two critical limitations in the context of outdoor Gaussian representations: (1) precisely predicting incident directions is inherently challenging, and (2) these methods fail to account for shadows cast by unmodeled objects outside the view frustum, resulting in incomplete shadow handling. Alternatively, we propose an image-based visibility extraction strategy that operates independently of explicit 3D modeling. The core of our approach is the construction of a ambient-only Gaussian representation ignoring cast sunlight.

Accurate estimation of the sun visibility map is critical for isolating the sun shading. Existing methods~\cite{li2022neulighting, bi2024gs3} rely on ray-surface intersection detection, which may fail to capture shadows cast by objects outside the view frustum. In contrast, we propose a residual-driven, image-based approach that avoids explicit 3D modeling.
%
% The core of our approach is to train an ambient-only Gaussian model as the basic scene representation, excluding all the effects of direct sunlight, i.e. sun shading and shadows. Then all sun-related effects will just be the residual of the model predictions and the corresponding truths, from where we derive the visibility map by pixel-wise clustering.
%
Specifically, we first train an ambient-only Gaussian model, excluding direct sunlight effects, \ie sun shading. Each Gaussian is assigned a predicted ambient color $ {\bm c_k^{amb}} $, formulated as:
\begin{equation}
\label{Eqn:amb_color}
    {\bm c_k^{amb}} = {{\Phi}}_{amb}({\bm f^{amb}_k}, \ {{\mathcal E}_i^{amb}}),
\end{equation}
where $ {\Phi}_{*} $ is a lightweight multi-layer perceptron (MLP), $ \bm f_k^{amb} $ denotes per-Gaussian local features, and ${\mathcal E}_i^{amb}$ represents per-image ambient embeddings. The ambient-only representation is optimized via a masked L1 loss:
\begin{equation}
\label{Eqn:amb_loss}
    {\mathcal L}_{amb} = \| (\hat {\mathcal I}_{amb} - \mathcal I) \cdot (1 - {\hat {\mathcal V}}_c) \|_1,
\end{equation}
% where $ {\hat {\mathcal{I}}_{amb}} = \sum_{k \in K} \omega_k {\bm c}_k^{amb} $, $ {\hat {\mathcal{V}}_c} $ is the coarse sun visibility aims to shield sunlight interference. (Refer to the appendix for more details on $ {\hat {\mathcal{V}}_c} $). Then the two-class clustering algorithm is performed on the residuals $ |\hat {\mathcal{I}}_{amb} - \mathcal{I} | $ between the ground truth training image $ \mathcal{I} $, which contains global illumination, and the corresponding ambient-only representation $ \hat {\mathcal{I}}_{amb} $, generating an accurate sun visibility $\hat {\mathcal V} $.
with $\hat{\mathcal I}_{amb}$ as the rendered ambient image obtained by splatting ${\bm c_k^{amb}}$, $\mathcal I$ as the ground truth image with global illumination, and $\hat{\mathcal V}_c$ a coarse visibility map derived from preprocessing (see supplementary material for details).
Subsequently, the residual map ${\hat {\mathcal I}}_{res} = |\mathcal I - \hat{\mathcal I}_{amb}|$ is computed to highlight regions affected by direct sunlight and shadows. A binary clustering algorithm is then applied to ${\hat {\mathcal I}}_{res}$, yielding the final refined sun visibility map:
\begin{equation}
\label{Eqn:visibility_extraction}
    {\hat {\mathcal{V}}} = 
\begin{cases} 
        1, & \text{if } \|{\hat {\mathcal I}_{res}} - \mu_1 \|^2 < \|{\hat {\mathcal I}_{res}} - \mu_2 \|^2 \\
        0, & otherwise
\end{cases}
\text{,}
\end{equation}
where $\mu_1$ and $\mu_2$ are the iteratively updated cluster centroids. The resulting $\hat{\mathcal V}$ is assigned 1 to sunlit areas and 0 otherwise.

\begin{figure*}[th]
  \centering
  \includegraphics[width=1.0 \textwidth]{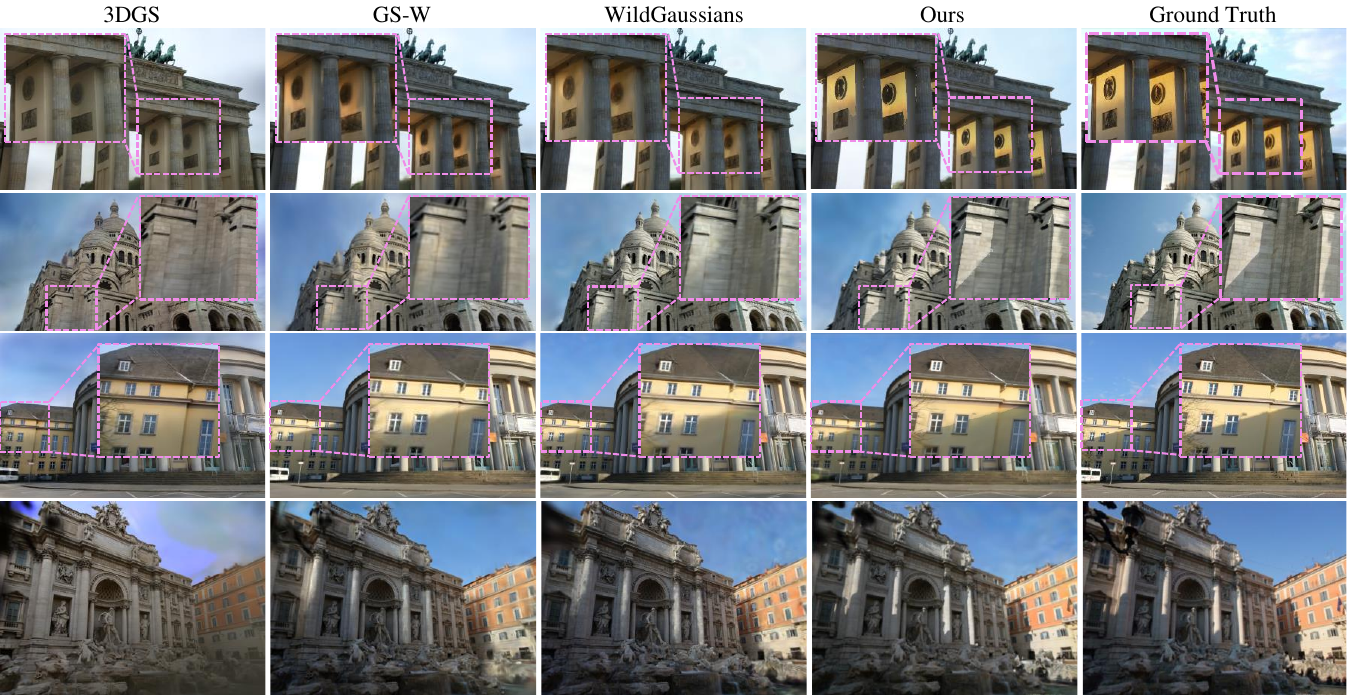}
  \caption{Comparison with SOTA methods in outdoor NVS. Our method delivers high-quality texture reconstruction for landmarks and surpasses existing methods in synthesizing sunlight, closely matching ground truth with sharp and accurate boundaries between illuminated and shadowed regions. In contrast, 3DGS~\cite{kerbl20233d}, GS-W~\cite{zhang2025gaussian}, and WildGaussians~\cite{kulhanek2024wildgaussians} suffer from severe blurring and distortion in light and shadow rendering.}
  \label{fig:nvs_compare}
\end{figure*}

\subsection{Illumination Decomposition}
\label{sec:illumination_decomposition} 
Inspired by intrinsic image decomposition (Eqn.~\ref{Eqn:intrinsic_image}),  for each Gaussian we predict the following spatial reflectance and shading colors (Fig.~\ref{fig:framework}):
\begin{flalign}
\label{Eqn:sun}
    & {\bm c}_k^{ref} = {\Phi_{ref}\Big({\bm f^{ref}_k}}\Big), \\
    & {\bm c}_k^{sun} = {{\Phi}}_{sun}\Big({\bm f^{sun}_k}, \ {{\mathcal E}_i^{sun}}\Big),\\
    & {\bm c}_k^{sky} = {{\Phi}}_{sky}\Big({\bm f^{sky}_k}, \ {{\mathcal E}_i^{sky}}\Big), \\
    & {\bm c}_k^{ind} = {{\Phi}}_{ind}\Big({\bm c_k}, \ {\bm f^{ind}_k}, \ {{\mathcal E}_i^{ind}}\Big).
\end{flalign}
Here, $ {\bm f^{*}_k} $ denotes per-Gaussian spatial shading-related features, while $ {{\mathcal E}_i^{*}} $ serves as the embeddings used to capture image-dependent illumination effects, including sunlight, sky illumination, and indirect shading. In our settings, to ensure physical interpretability, we make the spatial colors $ {\bm c}_k^{ref} $ independent of $ \mathcal E $ hence consistent across all viewpoints, while ${\bm c}_k^{sun}$ and $ {\bm c}_k^{sky} $ exhibit view-dependent consistency since the per-image embedding is broadcast to all spatial Gaussians. For decoding indirect color $ {\bm c}_k^{ind} $, we further incorporate the Gaussian’s base color $ \bm c_k $, derived from the spherical harmonic, to better capture local effects from scene lighting sources. Next, the rasterization process splats these spatial colors into corresponding rendered components: reflectance $ {\hat {\mathcal{R}}} $, sun shading $ {\hat {\mathcal{S}}}_{sun} $, sky shading $ {\hat {\mathcal{S}}}_{sky} $, and indirect shading $ {\hat {\mathcal{S}}}_{ind} $. The final rendering $ \hat {\mathcal{I}} $ is synthesized via Eqn.~\ref{Eqn:intrinsic_image}.

To supervise training, the ground truth image $\mathcal I$ is segmented into disjoint regions corresponding to the different shading components using binary masks. As we call this region-based supervision. The specific training losses are defined as: 
\begin{flalign}
\label{Eqn:sun_loss}
    & {\mathcal {L}_1^{sun}} =  {\lambda_1^{sun}} \|(\hat {\mathcal{I}} - {\mathcal {I}}) \cdot \hat {\mathcal{V}}\|_1, \\
    & {\mathcal {L}_1^{sky}} =  {\lambda_1^{sky}} \|(\hat {\mathcal {S}}_{sky} \cdot {\hat {\mathcal R}} - {\mathcal {I}}) \cdot {\mathcal M}_{sky}\|_1, \\
    & {\mathcal {L}_1^{ind}} =  {\lambda_1^{ind}} \|( ({\hat {\mathcal {S}}}_{sky} + {\hat {\mathcal {S}}}_{ind}) \cdot {\hat {\mathcal R}} - {\mathcal {I}}) \cdot {\mathcal M}_{ind}\|_1,
\end{flalign}
where $ \hat {\mathcal{V}} $ is the sun visibility map from Sec.~\ref{sec:visibility_extraction}, indicating that sunlight influences only visible regions. $ {\mathcal M}_{sky} $ is the sky mask, ensuring that sky radiance solely determines the sky color for consistent relighting effects. The term $ {\mathcal M}_{ind} $ is defined as the union of the complements of $ \hat {\mathcal{V}} $ and $ {\mathcal M}_{sky} $, capturing indirect lighting that faithfully models secondary reflections. Please see supplementary materials for details about $ {\mathcal{M}}_{sky} $.

Region-based supervision inherently leaves some regions within each shading component under-constrained and causes structural inconsistencies, leading to unrealistic sunlight effects during relighting. To resolve this, we introduce a structural consistency loss (SCL) to propagate locally optimized colors across the entire image domain while tolerating hue variations:
\begin{equation}
\label{Eqn:diffusion_loss}
     {\text{SCL}}({\mathcal {S}}, {\mathcal R}, \mathcal{M}) = \| ({\nabla \mathcal S} - {\nabla \mathcal R}) \cdot \mathcal{M} \|_1,
\end{equation}
where $ \nabla $ denotes the first-order difference along the coordinate axes, and $ \mathcal M $ is a region mask. We impose the constraint that $ {\hat {\mathcal {S}}}_{sun} $, $ {\hat {\mathcal {S}}}_{sky} $ and $ {\hat {\mathcal {S}}}_{ind} $ share the same content structure as $ {\hat {\mathcal R}} $, since $ {\hat {\mathcal R}} $ consistently maintains a complete and well-defined content structure. This leads to the following loss function summed in total:
\begin{equation}
\label{Eqn:SC_loss}
    {\mathcal {L}_{sc}} = \sum_{i} {\lambda_{sc}^{i}} {\text {SCL}}({\hat {\mathcal {S}}}_{i}, \ {\hat {\mathcal R}}, \ 1 - \mathcal{M}_{sky}),
\end{equation}
where $ i \in \{ sun, sky, ind \} $ is the shading component.

\subsection{Ray-trace-based Shadow Modeling}
\label{sec:gaussian_visibility}
To model continuous shadow effects, we explore a ray-trace-based Gaussian visibility query method that adapts to arbitrary light directions. It takes the following steps:

\vspace{0.2cm}
\noindent {\bf Visibility Query by Ray-tracing.} 
For each Gaussian $ \mathcal{G}_k $, a ray is cast from its center along an arbitrary light direction $\bm d_i$. The ray-traced visibility is determined by the surplus transmittance after accounting for all occluding Gaussians during traversal. Specifically, the transmittance of each Gaussian along the ray is given by the following recursive formula:
\begin{equation}
\label{Eqn:vis_gaussian}
     T_j = (1 - \alpha_{j-1}) T_{j-1} {\bm n_{j-1}} {\bm d_i^\intercal}, \ \ {\text{with}} \ \ T_1=1,
\end{equation}
where $ {\bm n_j} $ is the Gaussian's normal approximated by its shortest principal axis, while its opacity $ \alpha_j $ reflects its contribution to the 3D Gaussian. After traversing, the remanent transmittance $ T_j $ corresponds to the ray-traced visibility $ v_k^{rt} $ of Gaussian $ \mathcal{G}_k $.

\vspace{0.2cm}
\noindent {\bf Two-step Filter Strategy.} Due to the geometric coarseness of Gaussian representations, inaccuracies may arise in the ray-tracing process. In particular, (1) floating Gaussians in the sky can obscure surface Gaussians, causing them to appear erroneously invisible; and (2) transient Gaussians from dynamic objects result in inconsistent query results across viewpoints. Thus, we propose a two-step filtering process—comprising the sky filter and front filter—prior to ray tracing. First, we project all Gaussians onto the image plane and remove those located in the sky region, as guided by the sky mask $ {\mathcal {M}}_{sky} $. Second, we discard Gaussians with depth values smaller than the corresponding rendered depth. This filtering procedure effectively refines the scene geometry and improves the accuracy of surface visibility, thereby yielding more realistic and coherent shadows.

\vspace{0.2cm}
\noindent {\bf Visibility Baking.} 
To reduce the computational overhead of ray tracing during inference, we bake the ray-traced visibility into per-Gaussian visibility features $ \bm f_{k}^{vis} $. Specifically, the predicted per-Gaussian visibility for a given ray direction $\bm d_i$ is formulated as:
\begin{equation}
\label{Eqn:visibility}
    {v_k} = \Phi_{vis}(f_k^{vis}, \ {\bm p_k}, \ {\bm d_i}),
\end{equation}
where $ \bm p_k $ (\ie the Gaussian center) is introduced to make the MLP spatial aware. For training, Fibonacci sampling is applied on the upper hemisphere to randomly generate sunlight incident directions $ \bm d_i $. For each sample, two sets of Gaussian visibility $ v_k^{rt} \in \mathbb{R}^+ $ and $ v_k \in [-1, 1] $ are simultaneously computed via ray tracing and the prediction from Eqn.~\ref{Eqn:visibility}. The corresponding rendered visibility maps $\hat{\mathcal V}_{rt}$ and $\hat{\mathcal V}$ are then aligned by:
\begin{equation}
\label{Eqn:normal_loss}
    {\mathcal {L}_{vis}} = \| (\hat {\mathcal{V}} - \hat {\mathcal{V}}_{rt}) \cdot (1 - {\mathcal {M}}_{sky})\|_2,
\end{equation}
where the term $(1 - {\mathcal M}_{sky})$ ensures that optimization focuses on regions outside the sky, while the sky region's visibility is fixed to 1, indicating that sunlight always reaches the sky. During relighting, the rendered visibility map $ \hat {\mathcal{V}} $ can directly generate shadow effects through the Eqn.~\ref{Eqn:intrinsic_image}. 

\begin{table*}[t]
    \caption{Quantitative comparison with NeRF- and 3DGS-based relighting methods on Staats Theater, Brandenburg Gate, Trevi Fountain, and Sacre Coeur datasets. The \colorbox{tabfirst} {first}, \colorbox{tabsecond} {second}, and \colorbox{tabthird} {third} best-performing methods are highlighted. All metrics are evaluated on one NVIDIA GeForce RTX 3090.}
    \label{tab:result}
    \begin{adjustbox}{max width=\linewidth}
    \setlength{\tabcolsep}{3mm}
    \centering
    {\fontsize{10}{15}\selectfont
    \begin{tabular}{l|ccc|ccc|ccc|ccc|c|c}
    \toprule
    \specialrule{0.05em}{0em}{0em}
    
    \multicolumn{1}{l|}{\multirow{2}*{Methods}} & 
    \multicolumn{3}{c|}{Staats Theater} & 
    \multicolumn{3}{c|}{Brandenburg Gate} & 
    \multicolumn{3}{c|}{Trevi Fountain} &
    \multicolumn{3}{c|}{Sacre Coeur} &
    \multicolumn{1}{l|}{\multirow{2}*{Train hrs.$\downarrow$}} &
    \multicolumn{1}{l}{\multirow{2}*{FPS$\uparrow$}} \\
    
    & PSNR$\uparrow$ & SSIM$\uparrow$ & LPIPS$\downarrow$ & PSNR$\uparrow$ & SSIM$\uparrow$ & LPIPS$\downarrow$ & PSNR$\uparrow$ & SSIM$\uparrow$ & LPIPS$\downarrow$ & PSNR$\uparrow$ & SSIM$\uparrow$ & LPIPS$\downarrow$ & & \\
    \hline 

    NeRF-W          & 19.63 & 0.662 & 0.388 & 24.17 & 0.891 & 0.167 & 18.97 & 0.700 & 0.265 & 19.20 & 0.808 & 0.192 & 156 & \textless1 \\
    NeRF-OSR        & 15.43 & 0.603 & 0.401 & 17.43 & 0.842 & 0.174 & 19.43 & 0.706 & 0.258 & 20.03 & 0.806 & 0.184 & 186 & \textless1 \\
    Ha-NeRF         & 21.03 & 0.671 & 0.382 & 24.04 & 0.887 & 0.139 & 20.18 & 0.691 & 0.223 & 20.02 & 0.801 & 0.181 & 436 & \textless1 \\
    3DGS            & 13.42 & 0.638 & 0.392 & 19.33 & 0.884 & \cellcolor{tabthird} 0.132 & 17.08 & 0.714 & 0.241 & 17.70 & 0.845 & 0.186 & \cellcolor{tabfirst} 2.6 & \cellcolor{tabsecond} 52 \\
    SWAG            & \cellcolor{tabthird} 22.29 & 0.713 & 0.364 & 26.33 & \cellcolor{tabthird} 0.929 & 0.139 & \cellcolor{tabthird} 23.10 & \cellcolor{tabfirst} 0.815 & \cellcolor{tabthird} 0.208 & 21.16 & \cellcolor{tabthird} 0.860 & 0.185 & \cellcolor{tabsecond} 3.6 & 13 \\
    WildGaussians   & \cellcolor{tabsecond} 22.58 & \cellcolor{tabthird} 0.761 & \cellcolor{tabthird} 0.346 & \cellcolor{tabthird} 27.77 & 0.927 & 0.133 & \cellcolor{tabsecond} 23.63 & 0.766 & 0.228 & \cellcolor{tabthird} 22.56 & 0.859 & \cellcolor{tabthird} 0.177 & 7.4 & \cellcolor{tabfirst} 108 \\
    GS-W            &21.69 & \cellcolor{tabsecond} 0.772 & \cellcolor{tabsecond} 0.344 & \cellcolor{tabsecond} 27.96 & \cellcolor{tabsecond} 0.932 & \cellcolor{tabfirst} 0.086 & 22.91 & \cellcolor{tabthird} 0.801 & \cellcolor{tabfirst} 0.156 & \cellcolor{tabfirst} 23.24 & \cellcolor{tabsecond} 0.863 & \cellcolor{tabfirst} 0.130 & 6.4 & \cellcolor{tabthird} 43\\
    \hline 
    {\bf Ours}      & \cellcolor{tabfirst} 22.84 & \cellcolor{tabfirst} 0.774 & \cellcolor{tabfirst} 0.243 & \cellcolor{tabfirst} 28.16 & \cellcolor{tabfirst} 0.938 & \cellcolor{tabsecond} 0.129 & \cellcolor{tabfirst} 23.84 & \cellcolor{tabsecond} 0.806 & \cellcolor{tabsecond} 0.201 & \cellcolor{tabsecond} 22.70 & \cellcolor{tabfirst} 0.875 & \cellcolor{tabsecond} 0.163 & \cellcolor{tabthird} 4.6 & 32\\
    \bottomrule
    \specialrule{0.05em}{0em}{0em}
\end{tabular}}
\end{adjustbox}
\end{table*}

\begin{figure*}[th]
  \centering
  \includegraphics[width=1.0 \textwidth]{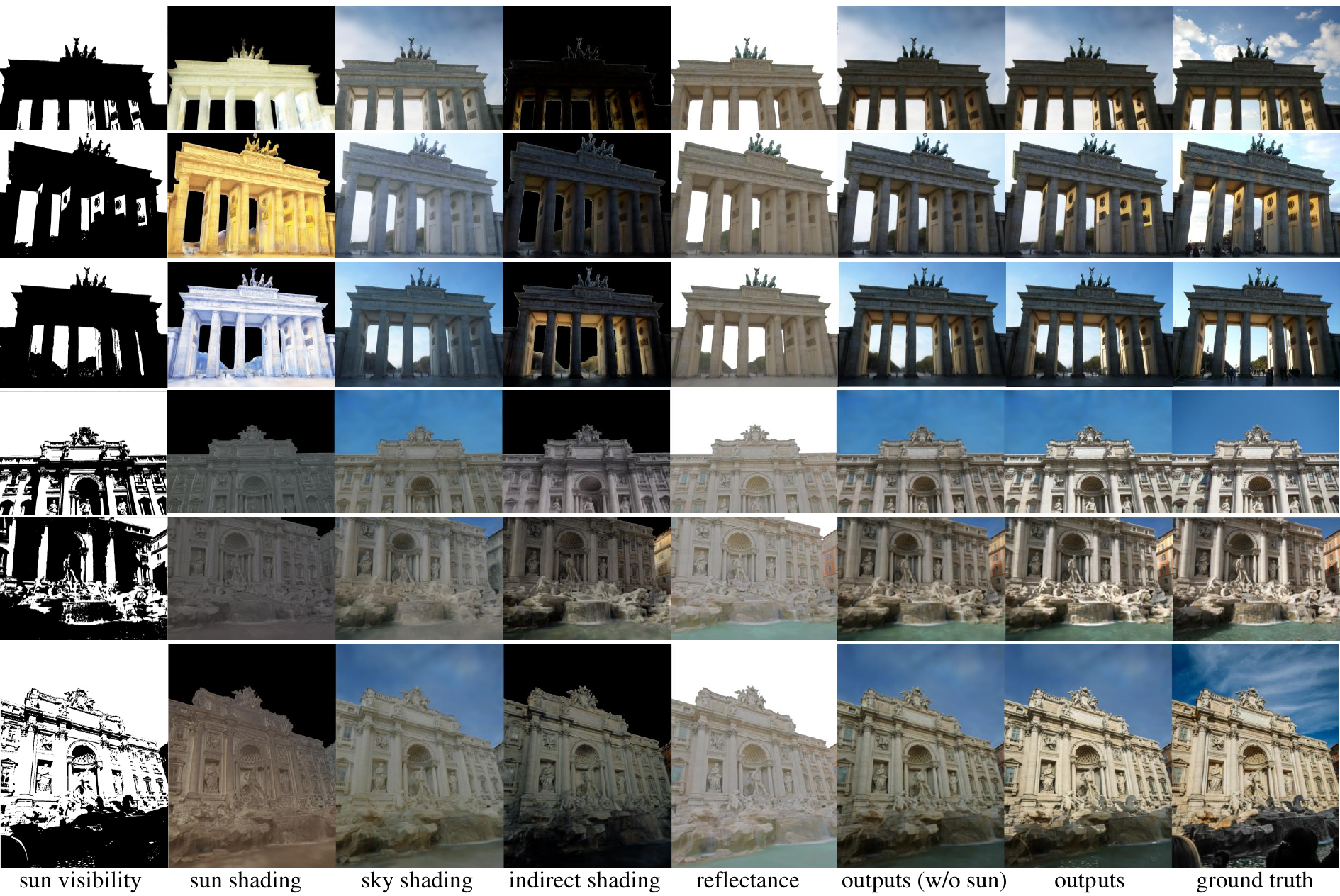}
  \caption{Visualization of the decomposed components and synthesis results. Sun shading captures the effects of direct sunlight, sky shading represents diffuse radiance from the sky, indirect shading models global illumination interactions, and reflectance preserves material properties independent of lighting. Combined with the sun visibility map, these components reconstruct the final illumination with high accuracy and physical plausibility.}
  \label{fig:illumination}
\end{figure*}

\begin{figure*}[t]
  \centering
  \includegraphics[width=1.0 \textwidth]{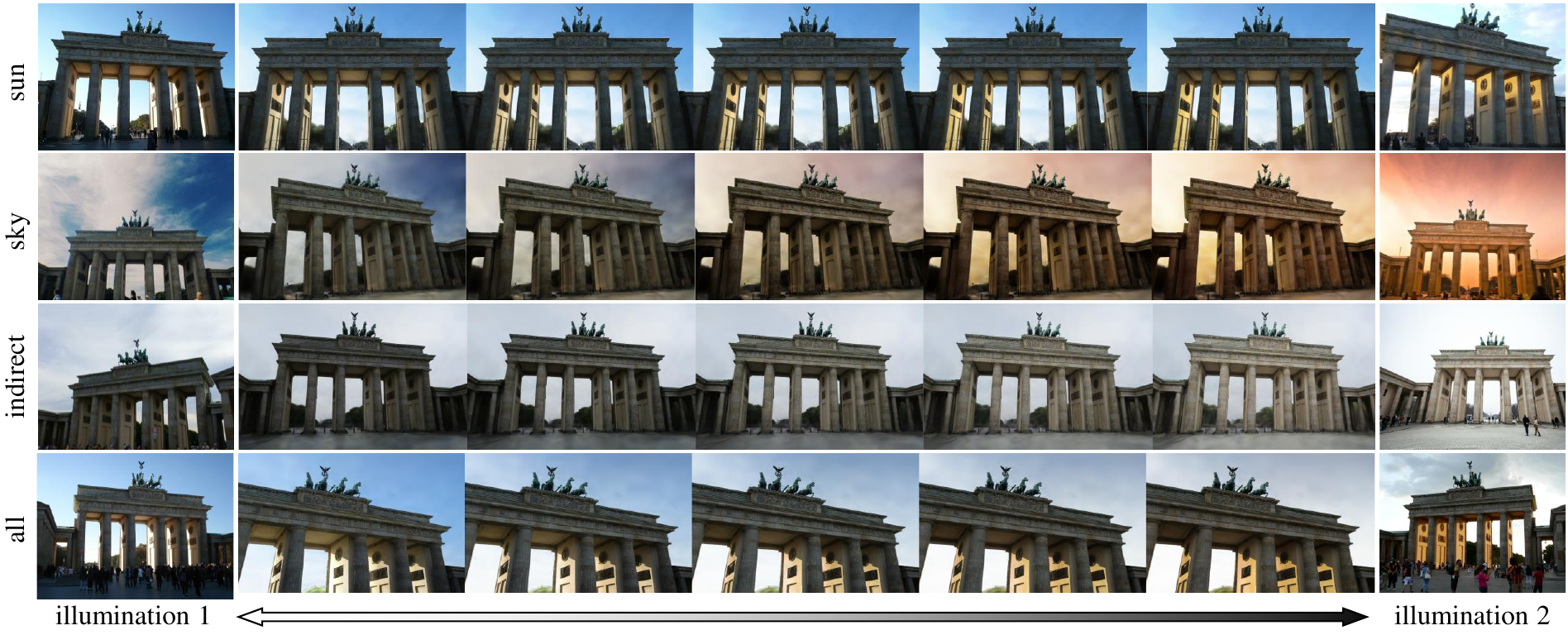}
  \caption{Relighting through illumination interpolation. Visualization shows how scene appearance evolves when interpolating between two arbitrary illumination conditions. Each shading component—sun shading, sky shading, and indirect shading—is interpolated both individually and collectively (All), highlighting the ability to control lighting variations at a granular level. The results showcase diverse relighting effects, with smooth transitions and sharp details, particularly in regions with high-frequency features like light-shadow boundaries. The combination of physically interpretable components ensures highly realistic and visually coherent relighting outcomes.}
  \label{fig:relighting}
\end{figure*}

% {\bf Sky semantic.} The above two-step filtering strategy effectively mitigates the misleading influence of floating Gaussians in the sky on surface visibility. However, it introduces inaccuracies in the visibility of the sky region. To resolve this, we define a sky semantic $ \bm o_k \in \mathbb{R}^+ $ for each Gaussian and transform the optimization as a binary classification task. This is done by by minimizing the cross-entropy loss ($CEL$) between the rendered sky mask $ { \hat {\mathcal {M}}}_{sky} $, generated by splatting the sky semantic, and the ground truth sky mask $ {\mathcal {M}}_{sky} $, obtained through dataset preprocessing (see supplementary materials for details). The loss function is:
% \begin{equation}
% \label{Eqn:semantic_loss}
%     {\mathcal {L}_{sem}} = CEL({\mathcal {M}}_{sky}, { \hat {\mathcal {M}}}_{sky}).
% \end{equation}

\vspace{0.2cm}
Notably, our framework seamlessly integrates diverse shading-related features $ \{ f_k^{ref}, f_k^{sun}, f_k^{sky}, f_k^{ind} \} $ and visibility features $ f_k^{vis} $ within a unified Gaussian representation, enabling the simultaneous synthesis of all shading components $ \{ {\hat {\mathcal{R}}}, {\hat {\mathcal{S}}}_{sun}, {\hat {\mathcal{S}}}_{sky}, {\hat {\mathcal{S}}}_{ind} \} $ and shadow conditions $ \hat {\mathcal{V}} $ in a single rasterization pass for real-time relighting.
 
\section{Implementation Details}
All networks in our framework are implemented as compact 3-layer MLPs with ReLU activation applied after each hidden layer; the visibility baking network uses layer sizes $ \{32, 32, 32 \} $ with a Tanh output, whereas the others use layer sizes $ \{64, 64, 64 \} $ with a Sigmoid output. The whole framework employs a three-stage training manner to optimize parameters: (1) purely ambient-only scene representation (Sec.~\ref{sec:visibility_extraction}), (2) shading components optimization (Sec.~\ref{sec:illumination_decomposition}), and (3) visibility baking (Sec.~\ref{sec:gaussian_visibility}), with training iterations set to 10k, 100k, and 20k, respectively. We initialize per-Gaussian shading features, visibility features, and embeddings with 18, 6, and 32 dimensions, respectively. During training, we use learning rates of 2.5e-3 for shading features and 1.0e-3 for visibility features, while the MLPs and embeddings begin with learning rates of 8.0e-4 and 5.0e-3, respectively, decaying exponentially to 5.0e-6 and 5.0e-5. Regularization parameters are set as $ \lambda_1^{sun}=1.0 $, $ \lambda_1^{sky}= \lambda_1^{ind}=10.0 $, $ \lambda_{sc}^{sun}=0.1 $, and $ \lambda_{sc}^{sky}=\lambda_{sc}^{ind}=5.0 $.

\section{Results Analysis}

\subsection{Performance on Novel View Synthesis}
We evaluate the performance of novel view synthesis (NVS) on four outdoor scenes: Staats Theater and Trevi Fountain from \cite{rudnev2022nerf}, as well as Brandenburg Gate and Sacre Coeur from \cite{martin2021nerf}. For comparison, we include NeRF-based methods~\cite{martin2021nerf, rudnev2022nerf, chen2022hallucinated} and 3DGS-based relighting approaches \cite{kerbl20233d, dahmani2025swag, zhang2025gaussian, kulhanek2024wildgaussians}, with a focus on their performance in challenging unconstrained scenarios. The quantitative results in Tab.~\ref{tab:result} demonstrate that our method achieves performance comparable to state-of-the-art (SOTA) methods while surpassing them in PSNR and SSIM metrics across diverse scenarios. In Fig.~\ref{fig:nvs_compare}, compared methods often struggle to accurately reconstruct the colors of light and shadow, frequently introducing artifacts. In contrast, our method effectively captures realistic light and shadow effects, delivering artifact-free results with sharp, well-defined shadow boundaries and seamless transitions.

\begin{figure}[t]
  \centering
  \includegraphics[width=0.48 \textwidth]{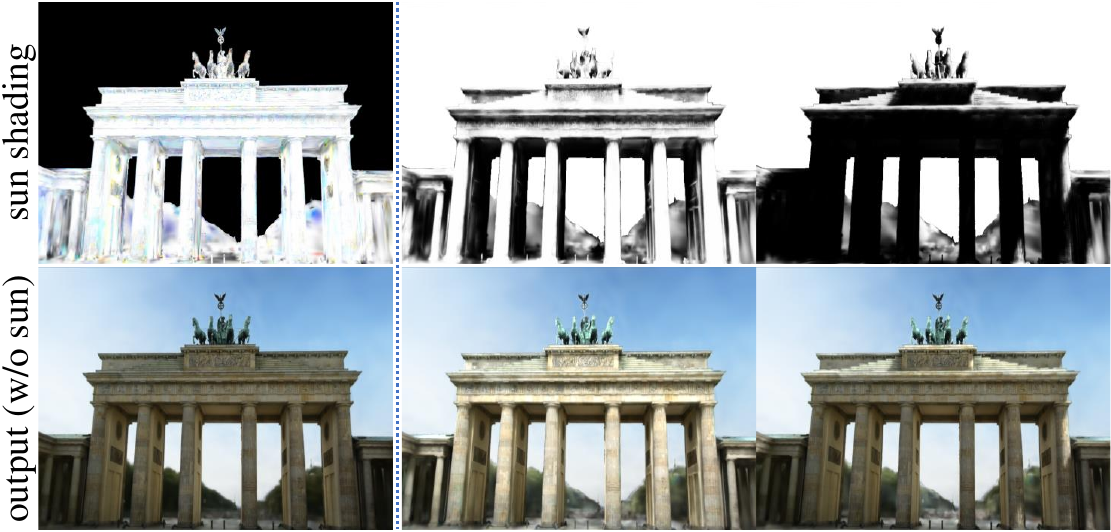}
  \caption{Visualization of shadow effect. 
  Left: (Top) Decomposed sun shading; (Bottom) Scene without sunlight. 
  Right: (Top row) Rendered sun visibility maps under different incident directions; (Bottom row) Final renderings with sunlight superimposed.}
  \label{fig:shadow}
\end{figure}

\subsection{Decomposition and Relighting}
{\bf Illumination Decomposition.} As shown in Fig.~\ref{fig:illumination}, the sun and sky shading components yield uniform diffuse reflection, while the indirect shading effectively captures localized variations from both direct and ambient light sources. This observation validates our approach for decoding spatial shading variations (Sec.~\ref{sec:illumination_decomposition}). Moreover, the sky color is exclusively determined by sky shading, ensuring that landmarks under sky illumination consistently reflect its hue. In contrast, the diffuse color and intensity of sunlight vary significantly with sun conditions. By leveraging accurate sun visibility, our synthesis achieves a precise, well-delineated integration of all shading components, with sunlight generating sharp, realistic shadow effects.

\noindent {\bf Appearance Interpolation.} 
By decomposing global illumination into multiple shading components beyond a single appearance embedding~\cite{kulhanek2024wildgaussians, martin2021nerf}, our method enhances the capabilities of lighting re-rendering, enabling the synthesis of more diverse and physically plausible relighting effects.
As demonstrated in Fig.~\ref{fig:relighting}, our approach enables the interpolation of shading embeddings from different reference views, allowing for the independent adjustment of specific shading components, such as direct sunlight, sky shading, and indirect illumination. Additionally, it supports the simultaneous modification of multiple components. This fine-grained control not only facilitates the reproduction of realistic lighting conditions but also enables the creation of novel and compelling illumination effects.

% By decomposing global illumination into multiple shading components beyond a single appearance embedding~\cite{kulhanek2024wildgaussians, martin2021nerf}, our method expands the scope of lighting re-rendering, enabling the generation of more diverse and realistic relighting effects. As shown in Fig.~\ref{fig:relighting}, our approach interpolates shading embeddings from different reference views, allowing for the independent adjustment of specific shading components—such as direct sunlight, sky shading, or indirect illumination—as well as the simultaneous modification of multiple components to achieve complex relighting scenarios. Notably, since each shading component contributes exclusively to a specific local region in the final synthesis, our method enhances flexibility in relighting control.

\noindent {\bf Shadow Effect.} 
A key contribution of our relighting framework is its ability to synthesize consistent shadows that capture realistic light–shadow dynamics. 
As depicted in Fig.~\ref{fig:shadow}, rendered sun visibility maps (top row, right) reveal that the illuminated regions of the scene can adapt dynamically to variations in the incident light direction. By conditioning decomposed sun shading on accurately computed sun visibility, the sun shading (top left) is seamlessly integrated with the unlit scene (bottom left), resulting in a natural and precise interplay between light and shadow (bottom row, right).

\begin{figure}[t]
  \centering
  \includegraphics[width=0.48 \textwidth]{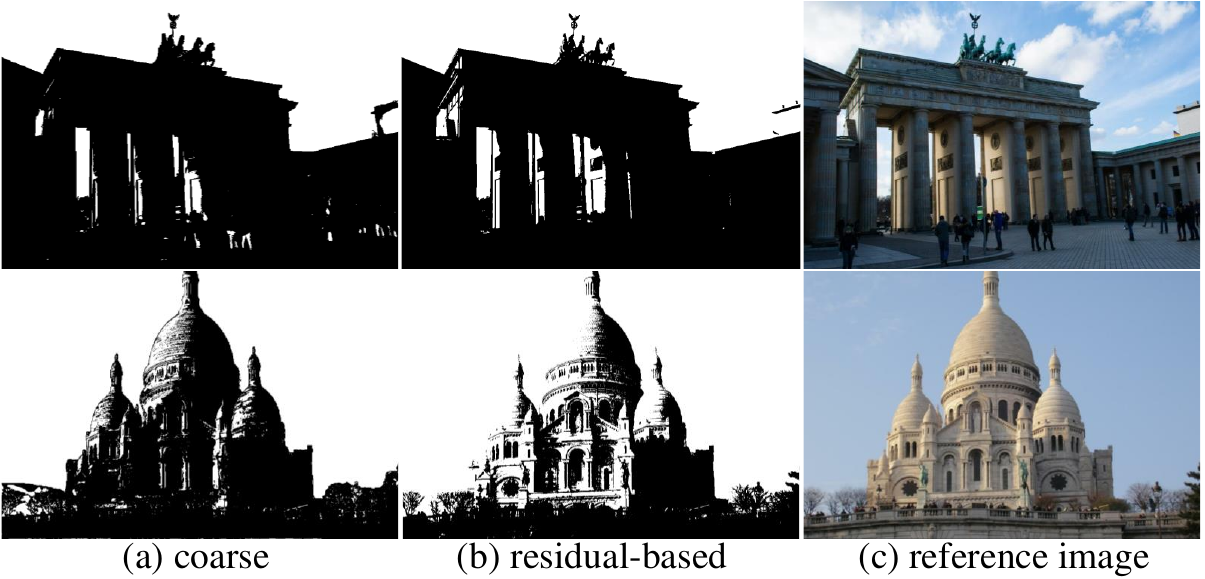}
  \caption{Comparison of visibility extraction methods. (a) The coarse visibility derived from dataset preprocessing; (b) The visibility map obtained using our residual-based method.}
  \label{fig:refine_vis}
\end{figure}

\begin{figure}[t]
  \centering
  \includegraphics[width=0.48 \textwidth]{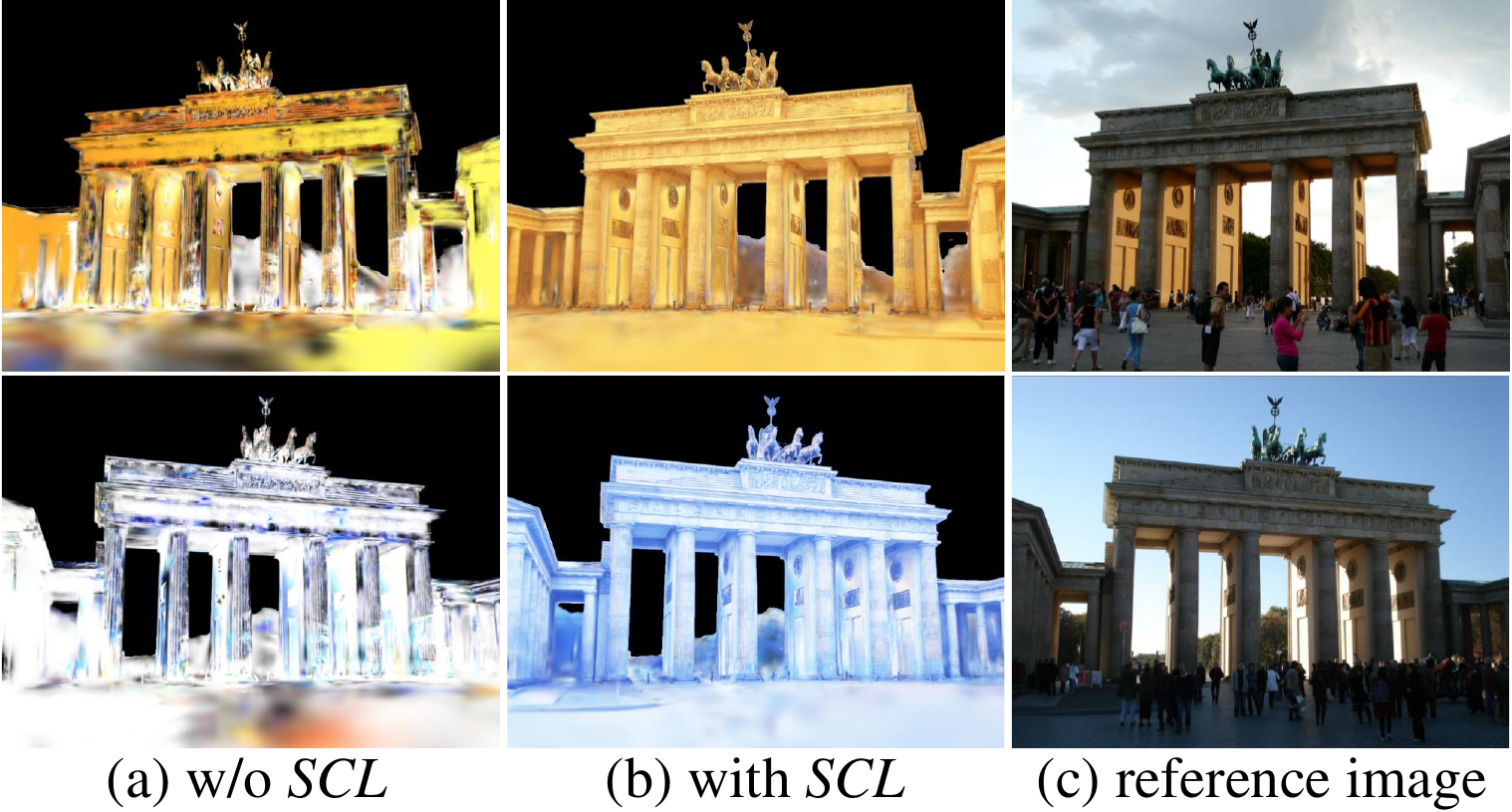}
  \caption{Comparison of the effect of structural consistency loss (SCL) on sun shading. (a) Without SCL, sun shading is well-optimized within directly lit regions but becomes under-constrained outside these areas. In contrast, (b) applying SCL enforces global consistency across the image domain.}
  \label{fig:SCL}
\end{figure}

\begin{figure}[t]
  \centering
  \includegraphics[width=0.48 \textwidth]{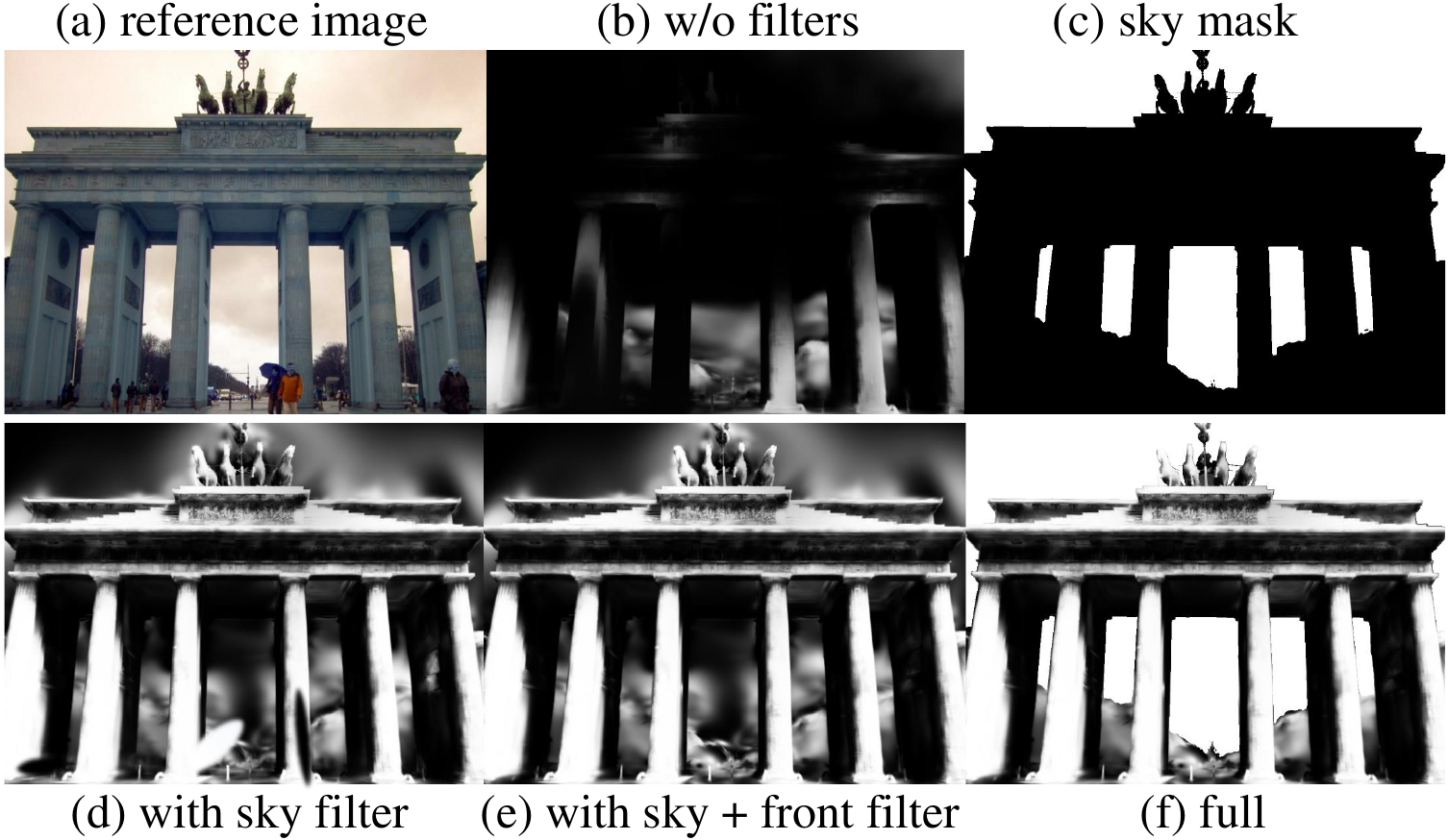}
  \caption{Ablation of two-step filter strategy. (a) Reference view; (b) initial result from direct visibility query through ray tracing; (c) sky mask; (d) sky filter process suppressing interference from floating Gaussians; (e) front filter resolves ambiguities from transient Gaussians; (f) Final visibility map guided by sky mask.}
  \label{fig:filters}
\end{figure}

\subsection{Ablations}
{\bf Residual-based Visibility Extraction.} 
The coarse visibility map (Fig.~\ref{fig:refine_vis}(a)) frequently exhibits incomplete coverage of sunlit regions, particularly near intricate boundaries and areas with subtle transitions. In contrast, our residual-based method produces a refined visibility map (Fig.~\ref{fig:refine_vis}(b)) that aligns almost perfectly with the sunlit regions in the globally illuminated image (Fig.~\ref{fig:refine_vis}(c)). This demonstrates the residual-based method’s superior capability in extracting precise and comprehensive visibility maps, which are essential for accurate sun shading isolation.

\noindent {\bf Efficacy of Structural Consistency Loss ($\textbf {SCL}$).} 
The objective of SCL is to mitigate the global under-constraint in shading that results from exclusive reliance on region-based supervision. For instance, in the case of sun shading (Fig.~\ref{fig:SCL}(a)), proper optimization is achieved in the sunlit areas while other regions exhibit significant uncertainty. The introduction of SCL effectively enforces global illumination consistency, as clearly demonstrated in Fig.~\ref{fig:SCL}(b).

\noindent {\bf Efficacy of Two-step Filter Strategy.} 
As shown in Fig.~\ref{fig:filters}(b), direct ray-tracing queries for Gaussian visibility often result in inaccurate shadow maps. The proposed sky filter effectively suppresses this interference, ensuring plausible visibility for scene surfaces (Fig.~\ref{fig:filters}(d)). Furthermore, the depth-guided filter resolves ambiguities caused by transient Gaussians in front of the geometric surface (Fig.~\ref{fig:filters}(e)). Incorporating sky mask compensates for the sky’s inherent visibility under arbitrary lighting directions, guaranteeing that the sky remains illuminated by the sun (Fig.~\ref{fig:filters}(c)). Together, these strategies produce a rendered visibility map that is both physically accurate and robust (Fig.~\ref{fig:filters}(f)).

\section{Conclusions}
We propose a Gaussian-based framework for outdoor relighting that integrates multiple shading effects from unconstrained photo collections. Our framework decomposes global illumination into consistent, physically interpretable shading components, enabling realistic appearance variations through independent or collaborative interpolation across diverse illumination conditions. Additionally, we incorporate ray tracing for Gaussian visibility queries, enabling shadow rendering on volume-based Gaussian representations. Furthermore, we integrate the ray-traced results into Gaussian features, allowing ray-direction adaptive shadow rendering. This approach computes all shading and shadow components in a single forward pass, enabling dynamic lighting and shadow effects for real-time relighting.

\section{Acknowledgement}
This work was supported by the National Natural Science Foundation of China (No.62032011), the Fundamental Research Funds for the Central Universities (0202-14380135), and the Nanjing University Technology Innovation Fund.

{
    \small
    \bibliographystyle{ieeenat_fullname}
    \bibliography{main}
}

% \newpage

\section{Dataset Preprocess}
Similar to the original Gaussian model, our method utilizes a dataset of pose-calibrated outdoor images and sparse point clouds generated via Structure-from-Motion (SfM). Additionally, we enhance the preprocessing step by extracting sky masks $ {\mathcal M}_{sky} $ and coarse sun visibility $ {\mathcal V}_c $ for all training images, which are used in Sec. 3.2 and Sec. 3.3. The detailed preprocessing steps are outlined below.

\subsection{Sky Mask} 
\label{sky_mask}
The sky mask is a binary mask that represents the sky and non-sky regions. To generate an accurate sky mask, we leverage the pre-trained Depth Anything V2~\cite{yang2024depth} model to estimate the depth maps for all photo collections, with predictions encoded as relative depth disparities. Given the empirical assumption that the sky lies at an infinite distance, it is characterized by near-zero disparity values. Thus, we adopt a disparity threshold of $ \tau_D = 0.1$, categorizing pixels as sky (1) or non-sky (0).

\subsection{Coarse Sun Visibility} 
Sun visibility is represented as a binary mask indicating whether sunlight can directly reach the surface of the scene. To extract sun visibility, we first pre-classify all photo collections into sunny and cloudy weather conditions based on~\cite{yu2021hierarchical}. Sun visibility extraction is performed only on sunny images, while for cloudy images, sun visibility is manually set to zero, indicating complete occlusion of the sun. For the sunny images, we convert the RGB images $ \mathcal{I} $ into the HSV color space, where the V channel $ \mathcal{I}_v $ represents brightness.  Sunlit regions exhibit higher brightness values, while shaded areas have lower values. To enhance the contrast between sunlit and non-sunlit regions, we apply a gamma transformation as follows:
\begin{equation}
\label{Eqn:gamma}
    {\mathcal {I}_V'} = { (\beta} \cdot{\mathcal{I}_V} - \epsilon)^{\gamma}.
\end{equation}
Here, $ \beta $, $ \epsilon $, and $ \gamma $ are the scaling coefficient, offset coefficient, and gamma transformation exponent, respectively, with values set to $ \frac{1}{255} $, 0.1 and 1.5. Subsequently, we apply a brightness threshold of $ \tau_V = 0.3 $  on $ \mathcal{I}_v' $ to classify the regions into two categories: areas with higher brightness values are labeled as sun-visible, while those with lower values are marked as sun-invisible. 

\section{Sky Semantics}
The two-step filtering strategy effectively mitigates the misleading influence of floating Gaussians in the sky on surface visibility. However, it introduces inaccuracies in the visibility of the sky region. To resolve this, we define a sky semantic $ \bm o_k \in \mathbb{R}^+ $ for each Gaussian and transform the optimization as a binary classification task. This is done by by minimizing the cross-entropy loss (CEL) between the rendered sky mask $ { \hat {\mathcal {M}}}_{sky} $, generated by splatting the sky semantic, and the ground truth sky mask $ {\mathcal {M}}_{sky} $, obtained through Sec.~\ref{sky_mask}. The loss function is:
\begin{equation}
\label{Eqn:semantic_loss}
    {\mathcal {L}_{sem}} = \text{CEL}({\mathcal {M}}_{sky}, { \hat {\mathcal {M}}}_{sky}).
\end{equation}
Once the sky semantics are optimized, the framework can render the sky mask $ { \hat {\mathcal {M}}}_{sky} $ from any arbitrary viewpoint. This capability facilitates continuous guidance for defining the sky region that remains consistently visible to the sun.

\section{Cloudy Day}
\begin{figure*}[t]
  \centering
  \includegraphics[width=1.0 \textwidth]{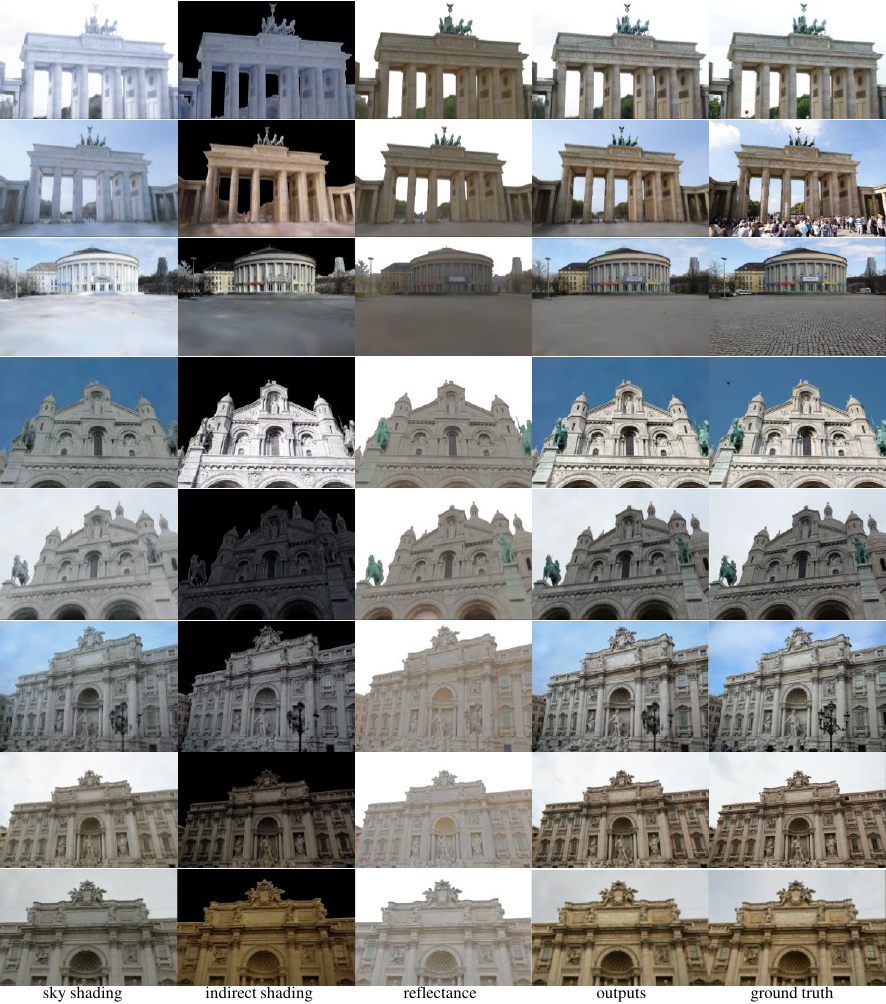}
  \caption{Visualization of the illumination decomposition results under cloudy conditions, where direct sunlight is absent. The scene is divided into sky shading, indirect shading, and reflectance. The sky shading models diffuse radiance from the overcast sky, while indirect shading captures global light transport within the scene. The reflectance preserves material properties independent of lighting. Together, these components enable a physically consistent and accurate reconstruction of the overall illumination in cloudy conditions.}
  \label{fig:cloudy}
\end{figure*}

In our experimental dataset, there are several instances representing scenarios without direct sunlight, $i.e.$, cloudy days. In these cases, the effect of sunlight is disregarded, and the global illumination for each viewpoint is decomposed into sky shading and indirect shading, as shown in Fig.~\ref{fig:cloudy}.

\section{Limitations and Future Work}
Our method is not without limitations. Firstly, due to the phased training process in our framework, the overall training duration is relatively longer compared to state-of-the-art methods such as 3DGS~\cite{kerbl20233d} and SWAG~\cite{dahmani2025swag}. However, it remains superior to Wildgaussians~\cite{kulhanek2024wildgaussians} and GS-W~\cite{zhang2025gaussian}, as demonstrated in Table 1. Second, achieving realistic shadow effects is highly dependent on accurate scene geometry. Although our approach effectively filters out Gaussians with ambiguous occlusion relationships in the sky and removes redundant Gaussians caused by transient objects, challenges remain in constructing precise scene surfaces. This is particularly true for areas like the ground, where the lack of texture leads to sparse Gaussian distributions, complicating the accurate extraction of sharp shadow contours. Lastly, our current framework is unable to transfer illumination from unseen images, which is a promising direction for future research.

\clearpage


\end{document}